%% file: Main_GC2024.tex
\documentclass[conference,10pt]{./MetaFiles/IEEEtran}

\IEEEoverridecommandlockouts

\input{./MetaFiles/packages}

\usepackage{amsmath}
\usepackage{algorithm}

\algrenewcommand\algorithmicforall{\textbf{foreach}}
\algrenewcommand\algorithmicindent{.8em}

\usepackage{pgfplots}
\usetikzlibrary{pgfplots.groupplots}

\usepackage{mathtools}

\begin{document}

\title{Adaptive Quantization Resolution and Power Control for Federated Learning over \\ Cell-free Networks 
}

\author{Afsaneh Mahmoudi, and Emil Björnson \\
 School of Electrical Engineering and Computer Science, KTH Royal Institute of Technology, Stockholm, Sweden \\
Emails: \{afmb, emilbjo\}@kth.se 
\thanks{The work was supported by the SUCCESS grant from the Swedish Foundation for Strategic Research.}
} 

\input{./MetaFiles/commands}
\maketitle

\begin{abstract}

Federated learning (FL) is a distributed learning framework where users train a global model by exchanging local model updates with a server instead of raw datasets, preserving data privacy and reducing communication overhead. However, the latency grows with the number of users and the model size, impeding the successful FL over traditional wireless networks with orthogonal access. Cell-free massive multiple-input multiple-output (CFmMIMO) is a promising solution to serve numerous users on the same time/frequency resource with similar rates. This architecture greatly reduces uplink latency through spatial multiplexing but does not take application characteristics into account.
In this paper, we co-optimize the physical layer with the FL application to mitigate the straggler effect. We introduce a novel adaptive mixed-resolution quantization scheme of the local gradient vector updates, where only the most essential entries are given high resolution. Thereafter, we propose a dynamic uplink power control scheme to manage the varying user rates and mitigate the straggler effect. The numerical results demonstrate that the proposed method achieves test accuracy comparable to classic FL while reducing communication overhead by at least $93\%$ on the CIFAR-10, CIFAR-100, and Fashion-MNIST datasets. We compare our methods against AQUILA, Top-$q$, and LAQ, using the max-sum rate and Dinkelbach power control schemes. Our approach reduces the communication overhead by $75\%$ and achieves $10$\% higher test accuracy than these benchmarks within a constrained total latency budget.


\end{abstract}





\section{Introduction}
Federated Learning (FL) enables the users in a wireless network to collaboratively train a machine learning model without sharing local datasets, enhancing data privacy and reducing communication overhead by transmitting model updates rather than raw data~\cite{konevcny2016federated}. In each iteration, users train local models on their private datasets and send updates to a central server, which aggregates and broadcasts the updated global model. Efficient communication networks are vital for exchanging high-dimensional local models rapidly, which is crucial for a successful FL. Cell-free massive multiple-input multiple-output (CFmMIMO) networks are well-suited for FL, offering the necessary uniformly high quality of service for uplink transmission of same-sized models from many users~\cite{cellfreebook}. These networks consist of numerous spatially distributed access points (APs) that provide coordinated service, leveraging macro-diversity and channel hardening to ensure consistent data rates despite imperfect channel knowledge.

FL reduces data communication overhead by avoiding data sharing, but it has substantial resource demands for transmitting high-dimensional gradient vectors in large-scale FL, which can still impede the training progress~\cite{9311931}. Therefore, efficient resource allocation—such as time, frequency, space, and energy—is essential for successful FL training. 


%
Adaptive quantization has gained significant attention in the domain of large-scale FL training. For instance, \cite{AQG} introduces an adaptive quantized gradient method that reduces communication costs while preserving convergence. Moreover,~\cite{AdaQuantFL} introduces AdaQuantFL, an adaptive quantization strategy for communication efficiency and low error, by changing the number of quantization levels during the training. The authors in \cite{JCDO} propose to accelerate Federated Edge Learning (FEEL) by data compression and setting deadlines to exclude stragglers, optimizing compression ratios and deadlines to minimize training time. Furthermore, \cite{FedDQ} offers a descending quantization scheme to reduce communication. AQUILA~\cite{AQUILA} presents an adaptive quantization approach to improve communication efficiency while assuring model convergence. A distributed approach for the communication of gradients called lazily aggregated quantized gradients~(LAQ) was proposed in \cite{LAQ}, while A-LAQ~\cite{ALAQ} introduces an adaptive-bit-allocation LAQ to reduce the FL communication overhead further. Moreover, \cite{DAdaQuant} presents DAdaQuant, a robust algorithm for dynamic adjustment of quantization levels, minimizing communication while maintaining convergence speed.

 To reduce the FL latency, \cite{9796621, 9124715} address joint optimization of user selection, transmit power, data rate, and processing frequency. The authors in \cite{9500541} optimize transmit power and data rate to minimize uplink latency in CFmMIMO-supported FL systems. The authors of~\cite{vu2021does} propose to allocate power and processing frequency in a CFmMIMO scheme for multiple FL groups to reduce FL iteration latency. Additionally,~\cite{FedAQ} uses adaptive quantization to reduce the uplink and downlink communication overhead. 

 The studies above have attempted to reduce the FL communication overhead by changing the quantization grid adaptively or optimal resource allocation, such as bits, powers, or frequencies. However, they applied the same quantization to the entire local gradient vector and did not exploit the sparsity of local gradients~\cite{SAFARI, 8849334, VectorQuantizedCompressedSensing}, often resulting in many near-zero elements. In this paper, we propose a novel adaptive element-wise quantization scheme for the local gradient vectors, where small entries are quantized with a single bit (representing the sign), and higher-value elements are uniformly quantized using multiple bits. 
 This approach massively reduces the required bits and makes it vary between users and iterations. We integrate the proposed solution into a CFmMIMO network by adapting the uplink powers to the varying number of bits and to mitigate the straggler effect. We compare the adaptive mixed-resolution quantization scheme numerically with AQUILA~\cite{AQUILA}, LAQ~\cite{LAQ}, and Top-$q$~\cite{wangni2018gradient} methods while applying our power control, Dinkelbach~\cite{EE_dinkel} and max-sum rate~\cite{cellfreebook}. Our proposed approach increases the test accuracy by~$10 \%$ while reducing the communication overhead by at least~$75\%$. 
 To the best of our knowledge, this is the first work that considers adaptive mixed-resolution quantization combined with reducing straggler effect in CFmMIMO.

\emph{Notation:} Italic font $w$, boldface lower-case $\bw$, boldface upper-case $\bW$, and calligraphic font $\calW$ denote scalars, vectors, matrices, and sets, respectively. We define the index set $[N] = \{1,2,\ldots,N\}$ for any positive integer $N$. We denote the $l_2$-norm by $\|\cdot\|$, the cardinality of the set $\calA$ by $|\calA|$, the entry $i$ of the vector $\bw$ by $[\bw]_i$, the entry $i,j$ of the matrix $\bW$ by $[\bW]_{i,j}$, and the transpose of $\bw$ by $\bw^\top$. 
\begin{figure}[t]
\centering
 \includegraphics[width=0.96\columnwidth] {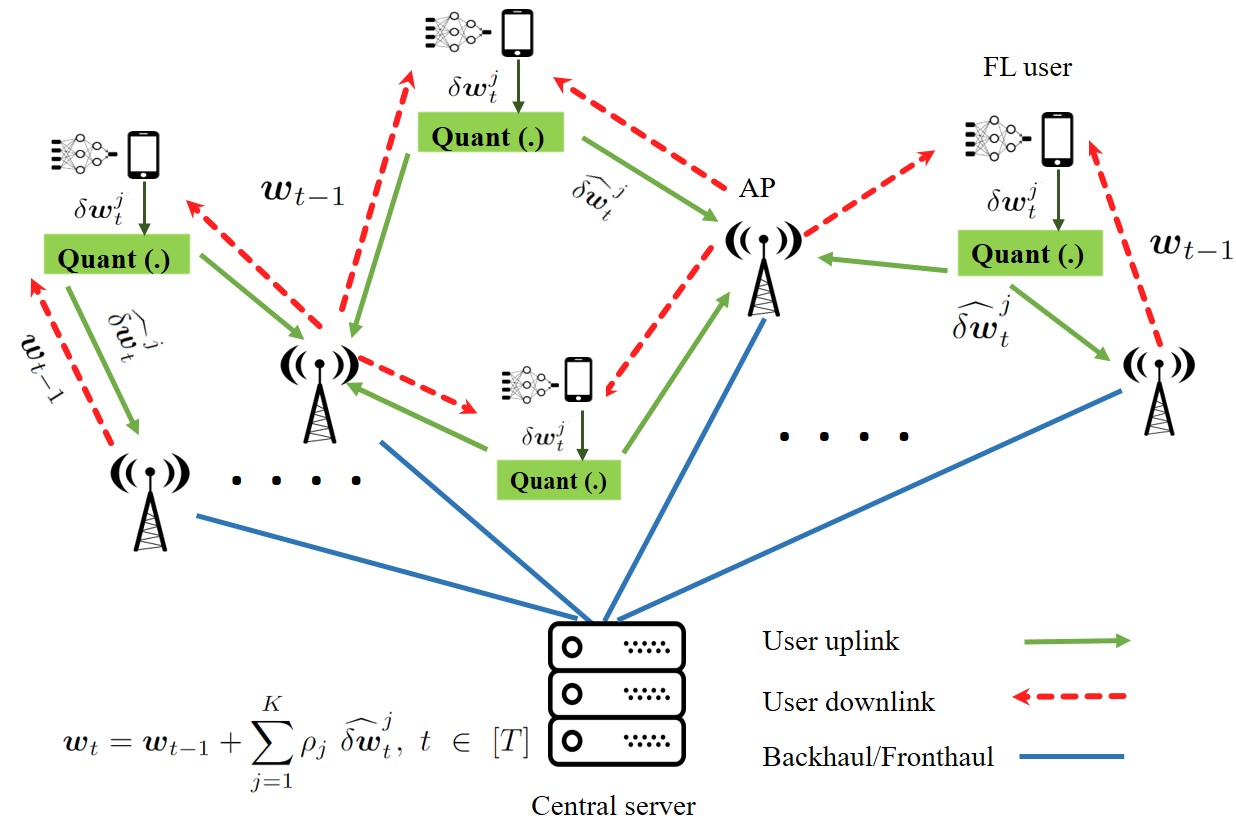}
  \caption{General architecture of FL over CFmMIMO with local quantization. }\label{fig: System_model_FLCFmMIMO}
\vspace{-0.02315\textheight} 
\end{figure}
 

\section{System Model and Problem Formulation}\label{section: System Model and Problem Formulation}
This section describes the FL setup, proposes a quantization scheme, presents the system model, and formulates the power control problem to reduce the straggler effect.

\subsection{Federated Averaging with Local AdaGrad Updates }\label{subsec: FLs}
We consider a setup where $K$ FL users cooperatively solve a distributed learning problem involving a loss function $f(\bw)$. The dataset $\calD$ is distributed among the users. The disjoint subset
$\calD_j$ is available at user $j \in [K]$ and satisfies $\calD_j \cap \calD_{j'}=\emptyset$ for $j \neq j'$ and $\sum_{j =1}^{K} |\calD_j| = |\calD|$.
We let the tuple $(\bx_{ij}, y_{ij})$ denote data sample $i$ of the $|\calD_j|$ samples available at user $j$.
We let $\bw \in \R^d$ denote the global model parameter with dimension~$d$ and define $\rho_j := {|\calD_j|}/{|\calD|}$ as the fraction of data available at user $j$. We formulate the training problem
\vspace{-0.005\textheight}
\begin{equation}\label{eq: w*}
\bw^* \in \argmin_{\bw\in\R^d} f(\bw)=\sum_{j =1}^{K} 
{\rho_j f_j(\bw)},
\vspace{-0.008\textheight}
\end{equation}
where $f_j(\bw) := \sum_{n =1}^{|\calD_j|} {f(\bw; \bx_{nj}, y_{nj})}/{|\calD_j|}$. 

To solve~\eqref{eq: w*}, we use federated averaging, an iterative algorithm with $T$ iterations. Initializing the training with~$\bw_0$, at the beginning of each iteration~$t \in [T]$, the server sends $\bw_{t-1}$ to the users. Then, every user $j\in[K]$ performs $L$ local iterations of AdaGrad updates similar to~\cite{reddi2020adaptive}. AdaGrad uses adaptive learning rates that adjust to gradient magnitudes, giving smaller gradients larger rates for balanced progress. This approach is particularly beneficial for deep neural networks, as it handles varying gradient scales and prevents large learning rates, enhancing convergence even with sparse gradients \cite{zeiler2012adadelta}. In the local updates, each user performs $i=1, \ldots, L$ local iterations with a randomly-chosen subset of $\xi_j \le |\mathcal{D}_j|$ data samples, and computes its local model~$\bw_{i,t}^j \in \mathbb{R}^d$, considering the initial points of $\bw_{0,t}^{j} = \bw_{t-1}$, and $\mathbf{g}_{1,t}^j = \mathbf{0}$:
\vspace{-0.01\textheight}
\begin{equation}\label{eq: local trains}
\begin{aligned}
  \quad \bw_{h,t}^{j} &\xleftarrow{} \bw_{h-1,t}^{j} - \frac{\alpha}{\sqrt{\mathbf{g}_{h,t}^{j} + \epsilon_a}} \odot \boldsymbol{\nabla}_{h-1,t}^{j},\,\, t\in[T], j\in[K], \\
  \quad \mathbf{g}_{h,t}^{j} &\xleftarrow{} \mathbf{g}_{h-1,t}^{j} +  \boldsymbol{\nabla}_{h-1,t}^{j} \odot \boldsymbol{\nabla}_{h-1,t}^{j},\,\, t\in[T],  j\in[K],
\end{aligned}
\end{equation}
where~$\boldsymbol{\nabla}_{h-1,t}^{j}: = \sum_{n\in[\xi_j]} \nabla_{\bw} f(\bw_{h-1,t}^{j}; \bx_{nj}, y_{nj})$, $\mathbf{g}_{h,t}^{j}$ is accumulated sum of squared gradients, $\odot$ denotes Hadamard product (element-wise multiplication), $\alpha$ is the preliminary step size, $\epsilon_a$ is a small constant to prevent division by zero and final local model is $\bw_t^j = \bw_{L,t}^{j}$. 
After computing~$\bw_t^j$, user $j$ calculates the local gradient~$\delta\bw_t^j:= \bw_t^j-\bw_{t-1}$, quantizes it as~$\widehat{\delta \bw}_t^j:= \text{Quant}(\delta\bw_t^j)$ and transmits~$\widehat{\delta \bw}_k^j$ to the central server over a CFmMIMO system with $M$ APs, which forward their received signals to the server over error-free fronthaul links, as shown in Fig.~\ref{fig: System_model_FLCFmMIMO}. We will propose a new quantization scheme in Section~\ref{subsec:sign-quant-scheme}. Based on the $K$ local gradients, the central server updates the global model as
  \vspace{-0.00\textheight}
 \begin{equation}\label{eq: iterative}
    \bw_{t} = \bw_{t-1} + \sum_{j =1}^{K}  \rho_j~\widehat{\delta \bw}_t^j,~t~\in~[T].
\end{equation}
Afterward, the central server sends the global model~$\bw_{t}$ to 
the APs via the fronthaul links. Next, the APs jointly send~$\bw_{t}$ to the users in the downlink. Finally, each user~$j \in [K]$ computes its new local model and begins the next FL iteration.

In this paper, we focus on the uplink transmission for sending quantized versions of the local gradient vectors to the APs by reducing the communication overhead with the proposed quantization scheme and reducing the straggler effect by uplink power control. The uplink is the weakest link because of the transmission of $K$ gradients from limited-powered devices. In contrast, the downlink transmits a single global model from the grid-connected APs with high power. Therefore, we assume the downlink is error-free. In the following subsection, we briefly explain the uplink process in CFmMIMO and refer to~\cite[Sec.~II-B]{WCNC} for further details.


\subsection{Uplink Process in FL over CFmMIMO}
\vspace{-0.0075\textheight}
We consider a CFmMIMO network consisting of $M$ APs, each with $N$ antennas and $K$ single-antenna FL users that will transmit their local vectors. We adopt a block-fading model where each channel realization spans $\tau_c$ channel uses~\cite{cellfreebook}. We assume independent Rayleigh fading for channels $g_{m,n}^{j} \sim \mathcal{N}_{\mathbb{C}}(0, \beta_{m}^{j})$ between user $j$ and AP $m$, with $\beta_{m}^{j}$ representing the large-scale fading coefficient. Each coherence block is divided into $\tau_p$ pilot channel uses and $\tau_c-\tau_p$ data channel uses. Each user transmits a $\tau_p$-length pilot $\sqrt{\tau_p} \boldsymbol{\varphi}_j \in \mathbb{C}^{\tau_p\times 1}$, using the pilot energy $p_p := \tau_p p^u$, where $p^u$ is the maximum transmit power. The received pilot signal at AP $m$ and antenna $n$ is given by $\by_{m,n}^{p} = \sqrt{p_p}\sum_{j=1}^K g_{m,n}^{j} \boldsymbol{\varphi}_j + \boldsymbol{\mu}_{m,n}^{p}$, where $\mu_{m,n}^p \sim \mathcal{N}_{\mathbb{C}}(0,{\color{black}\sigma^2})$ is the noise. In the uplink data transmission, each user $j$ transmits its quantized FL local gradient with power $p^u p_t^j$, where 
 $p_t^j \in [0,1]$ is the power control variable. After the APs receive the uplink signal, they use maximum-ratio combining. Similar to~\cite[Th.~2]{7827017}, the achievable uplink data rate (bit/s) at user $j$ and iteration~$t$ of the FL process is
\vspace{-0.01\textheight}
\begin{alignat}{3}\label{eq: Rk}
    &R_t^j = B_{\tau} \log_2\left(1+\text{SINR}_t^j\right),
    \vspace{-0.001\textheight}
\end{alignat}
where~${\text{SINR}}_t^j$ is the signal-to-noise+interference ratio, $B_\tau := B(1 - { \tau_p}/{\tau_c})$ is the pre-log factor, $B$ is the bandwidth~(in Hz),
\vspace{-0.015\textheight}
\begin{alignat}{3}
\nonumber
\text{SINR}_t^j &:= \frac{\bar{A}_j {p}_t^j}{\bar{B}_j {p}_t^j + \sum_{j' \neq j}^K p_t^{j'} \Tilde{B}_{j}^{j'} + I_M^j }, \bar{A}_j := \left( \sum_{m=1}^M N{\color{black}{\gamma_{m}^{j}}}\right)^{\! 2} \!\!,\: \\
      \nonumber
      I_M^{j} &:= \sum_{m=1}^M~N{\sigma^2{\gamma_{m}^{j}}}/{{p^u}},\quad \bar{B}_j := \sum_{m=1}^M N {\gamma_{m}^{j}} \beta_{m}^{j}, \: \\
      \Tilde{B}_{j}^{j'} &:= \sum_{m=1}^M N {\gamma_{m}^{j}} \beta_{m}^{j'} + |\boldsymbol{\varphi}_{j}^{H} \boldsymbol{\varphi}_{j'}|^2\left(\sum_{m=1}^M N {\gamma_{m}^{j}}\frac{\beta_{m}^{j'}}{\beta_{m}^{j}}\right)^2, \: \\
      \nonumber
        {\gamma_{m}^{j}}&:= \frac{p_p \left(\beta_{m}^{j}\right)^2}{p_p \sum_{j'=1}^{K} \beta_{m}^{j'} | \boldsymbol{\varphi}_{j'}^H \boldsymbol{\varphi}_j |^2 + \sigma^2}.
      \end{alignat}
In the following subsection, we describe the proposed adaptive mixed-resolution quantization scheme.
\vspace{-0.005\textheight}
\subsection{Adaptive Mixed-Resolution Quantization Scheme}
\label{subsec:sign-quant-scheme}
This subsection presents our proposed quantization scheme for uplink transmission of~$\delta \bw_t^j$, $j~\in~[K], t~\in~[T]$. Inspired by the inherent sparsity~\cite{SAFARI, 8849334, VectorQuantizedCompressedSensing} of the local gradients resulting in many near-zero elements (we call them low-resolution elements), we introduce an element-wise quantization approach that divides the elements into two quantization categories: low resolution and high resolution. User $j$ uses uniform quantization with $b_j \ge 2$ bits for the high-resolution elements obtained using a magnitude ratio threshold $\lambda_j$, while the remaining elements receive a single bit to represent their sign (0 or 1). This scheme is adaptive, as the number of high-resolution elements varies for each user~$j$ based on the values in each~$\delta \bw_t^j$. 
Next, for elements~$i=1,\ldots, d$, we define
\begin{equation}\label{eq: quant_factor}
     \bar{d}_t^j := \text{Number of elements}~i~\text{such that}~\frac{|[\delta\bw_t^j]_i|}{\|{\delta\bw_t^j}\|_{\infty}} \ge \lambda_j, 
\end{equation}
where~$\|\delta\bw_t^j\|_{\infty} := \max_{i\in [d]} |[\delta\bw_t^j]_i|$ is the infinity norm of the vector~$\delta \bw_t^j$. Let $\bb_t^j \in \R^{d \times 1}$ represent the bit vector, where the assigned bits for each element $i=1,\ldots,d$ of every $\delta\bw_t^j$, $j = 1,\ldots, K$, are as follows:
\begin{equation}\label{eq: bits}
 [\bb_t^j]_i =  
 \begin{cases}
     
     1, & \text{if~} \frac{|[\bw_t^j]_i|}{\nrm{\delta\bw_t^j}_{\infty}} < \lambda_j,~\text{and~} [\delta\bw_t^j]_i > 0,
     \\

     0, & \text{if~} \frac{|[\delta\bw_t^j]_i|}{\nrm{\delta\bw_t^j}_{\infty}} < \lambda_j,~\text{and~} [\delta\bw_t^j]_i \le 0, \\

     Q(|[\delta\bw_t^j]_i|), & \text{if~} \frac{|[\delta\bw_t^j]_i|}{\nrm{\delta\bw_t^j}_{\infty}} \ge \lambda_j,
\vspace{-0.01\textheight}
 \end{cases}
\end{equation}
where~$Q(\cdot)$ denotes the uniform quantization of the elements satisfying \eqref{eq: quant_factor} with $b_j$ bits (including one bit for the sign) and the grid radius of~$r_t^j:= \|{\delta \bw_t^j}\|_{\infty}-{\delta w}_{q,t}^{j}$, with ${\delta w}_{q,t}^{j}$ as the absolute value of the smallest element that satisfy \eqref{eq: quant_factor}. The bits $0$ and $1$ represent the signs of the elements that do not satisfy \eqref{eq: quant_factor}, with $0$ indicating a negative sign and $1$ indicating a positive sign. Defining~$s_t^j:=\bar{d}_t^j/d $, the total number of quantized bits is obtained as $b_t^j:= d ( b_j s_t^j + 1 - s_t^j ) + 32$ for user $j$ at iteration $t$, where $32$ bits are used to send~$r_t^j$. Note that all elements that do not satisfy \eqref{eq: quant_factor} have smaller absolute values than ${\delta w}_{q,t}^{j}$. 
Thus, the central server receives~$\widehat{\delta\bw}_t^j$ with elements~$i~\in[d]$ given by
\vspace{-0.015\textheight}
\begin{equation}\label{eq: w_hat}
    [\widehat{\delta\bw}_t^j]_i = 
    \begin{cases}
     
     {+\widehat{\delta w}_{q,t}^{j}}/{2}, &  \text{if}~[\bb_t^j]_i = 1, \\

     {-\widehat{\delta w}_{q,t}^{j}}/{2}, & \text{if}~[\bb_t^j]_i = 0, \\ 

     [\widehat{\delta \bw}_t^j]_i, & \text{otherwise},
 \end{cases}
 \vspace{-0.015\textheight}
\end{equation}
where~$\widehat{\delta w}_{q,t}^{j}$ is the quantized value of~${\delta w}_{q,t}^{j}$ with~$b_j$ bits. Since we apply one-bit uniform quantization with a radius grid of $|\widehat{\delta w}_{q,t}^{j}|$ for any element $i$ that does not satisfy \eqref{eq: quant_factor}, the quantized values for these elements are $\widehat{\delta w}_{q,t}^{j}$ divided by 2.
\vspace{-0.01\textheight}
\begin{lemma}\label{lemma: quant error}
   Let~$\widehat{\delta\bw}_t^j$ be the quantized local gradient vector corresponding to~$\delta \bw_t^j$ of user~$j$ at FL iteration~$t$. Let~$\boldsymbol{\varepsilon}_t^j:= \delta\bw_t^j - \widehat{\delta\bw}_t^j$ be the quantization error vector after quantizing~$\delta\bw_t^j$. For any~$j \in [K],~t \in [T]$, we obtain~$\|\boldsymbol{\varepsilon}_t^j \|_{\infty} \le c_j~\|\delta\bw_t^j \|_{\infty}$, where
   \vspace{-0.01\textheight}
   \begin{alignat}{3}\label{eq:  quant errors}
   & c_j:= \max \left\{  \frac{\lambda_j}{2} +\frac{ 1-\lambda_j}{4(2^{b_j}-1)},~\frac{ 1-\lambda_j}{2(2^{b_j}-1)}   \right\}.
   \end{alignat}
\end{lemma}
\vspace{-0.006\textheight}
\begin{IEEEproof}
    The proof is given in Appendix~A.
\end{IEEEproof}

Lemma~\ref{lemma: quant error} investigates the error in our mixed-resolution quantization method, showing that the quantization error decreases as the norm of the local gradient decreases. Building on this lemma, the following proposition demonstrates the convergence of FL with AdaGrad local updates.
\vspace{-0.005\textheight}
\begin{prop}\label{prop: convergence}
Let~$f_j(\bw)$ be the local loss function of user $j~\in [K]$ in the FL training with the local AdaGrad updates in~\eqref{eq: local trains}, $\widehat{\delta \bw}_t^j = {\delta \bw}_t^j - \varepsilon_t^j$ is the quantized local vector with the quantization error defined in~\eqref{eq:  quant errors}, and define~$c_{\max}:=\max_j~c_j$. Suppose $f_j(\bw)$ and $f_j(\bw)$ are~$\bar{L}$-smooth, for all $j~\in [K]$, with G-bounded gradients and $\sigma_l$-bounded (local) variance and the (global) variance $\sigma_g$ is bounded, similar to Assumptions~1-3 in~\cite{reddi2020adaptive}. Let~$\Delta_f:= f(\bw_0) - f(\bw^*)$, $\bar{\sigma}^2:=6 L \sigma_g^2 + \sigma_l^2$, $\alpha \le \min\{(16 L \bar{L})^{-1}, (2 L \sqrt{\bar{L}})^{-1} \}$, and the conditions~I-II in~\cite{reddi2020adaptive} hold. It then holds that
\vspace{-0.012\textheight}
\begin{alignat}{3}
    &\min_{t~\in~[T]}~\E_t  \nrm{\nabla f(\bw_t)}^2 \le \mathcal{O}\left(\frac{1 + \alpha L G \sqrt{T}  }{\alpha L T} \cdot  \Phi     \right), \: \\ 
 \nonumber 
 & \Phi:= 2L T \bar{\sigma}^2 + 4L^2 + K  c_{\max}^2 (2L + T \bar{\sigma}^2) + \alpha^{-1}\Delta_f.
\end{alignat} 
\end{prop}
\vspace{-0.008\textheight}
\begin{IEEEproof}
    The proof steps are similar to the proof of~[Th~1~in~\cite{reddi2020adaptive}] with the same assumptions. We substitute~$\Delta_t^j$ with~$\widehat{\delta \bw}_t^j$ containing the quantization errors, and assume~$\nu_t =0, \tau=1$. The details are omitted due to limited space.
\end{IEEEproof}

\subsection{Problem Formulation for Power Control}
\vspace{-0.005\textheight}
 Next, we present the proposed optimization problem that mitigates the straggler effect when using the proposed quantization scheme. Our goal is to obtain the uplink powers~$p_t^j$, $\forall j, t$, that minimize the latency of the slowest user:
 \vspace{-0.005\textheight}
 \begin{subequations}\label{eq: main optimization}
\begin{alignat}{3}
\label{optimization1}
  \underset{p_t^1, \ldots, p_t^K}{\mathrm{minimize}} & \quad \max_{j~\in~[K]} \quad \ell_t^j \: \\ 
  \text{subject to}
  \label{constraint1}
 &\quad 0 \le p_t^j \le 1,\quad \forall j~\in[K],~\forall t~\in[T],
\end{alignat}
\end{subequations}
where~$\ell_t^j$ is the uplink latency of each user~$j$ at iteration~$t$, as
\vspace{-0.006\textheight}
\begin{equation}\label{eq: ell-t-j}
    \ell_t^j := \frac{b_t^j}{R_t^j} = \frac{ d \left( b_j s_t^j + 1 - s_t^j\right) + 32 }{B_{\tau} \log_2\left(1+ {\text{SINR}}_t^j\right)}.
\end{equation}
We rewrite the min-max optimization problem in~\eqref{eq: main optimization} as
 \begin{subequations}\label{eq: equivalent}
\begin{alignat}{3}
\label{eqoptimization}
  \underset{p_t^1, \ldots, p_t^K}{\mathrm{maximize}} & \quad \min_{j \in [K]} \quad \frac{B_{\tau} \log_2(1 + {\text{SINR}}_t^j )}{b_t^j}  \: \\ 
  \text{subject to}
  \label{eqconstraint1}
 &\quad 0 \le p_t^j \le 1,\quad \forall j~\in[K],~\forall t~\in[T].
\end{alignat}
\end{subequations}
We can write \eqref{eq: equivalent} in epigraph form as 
\vspace{-0.01\textheight}
 \begin{subequations}\label{eq: equivalent1}
\begin{alignat}{3}
\label{eqoptimization1}
  \underset{p_t^1, \ldots, p_t^K,\eta_t}{\mathrm{maximize}} & \quad \eta_t  \: \\ 
  \text{subject to}
  \label{eqconstraint1}
 &\quad 0 \le p_t^j \le 1,\quad \forall j~\in[K],~\forall t~\in[T], \\ 
\label{eqconstraint2}
 & \quad  \left(\bar{A}_j - \theta_t^j \bar{B}_j \right)p_t^j - \theta_t^j \sum_{j' \neq j}^K p_t^{j'} \Tilde{B}_{j}^{j'} \ge \theta_t^j I_M^j.
\vspace{-0.01\textheight}
\end{alignat}
\end{subequations}
by letting the new optimization variable $\eta_t$ be a lower bound on the rate-per-bit ratio at FL iteration~$t$:
\vspace{-0.005\textheight}
\begin{equation}\label{eq: eta}
    \eta_t \leq \min_{j \in [K]} \quad \frac{B_{\tau} \log_2(1 + {\text{SINR}}_t^j )}{b_t^j}.
    \vspace{-0.008\textheight}
\end{equation}
We obtained \eqref{eqconstraint2} by expressing \eqref{eq: eta} as $\text{SINR}_t^j \ge 2^{\eta_t b_t^j/B_{\tau}}-1,~\forall~j,~t$ and then defining $\theta_t^j:= 2^{\eta_t b_t^j/B_{\tau}} - 1 \ge 0$.
 In the following section, we propose a way to solve~\eqref{eq: equivalent1} optimally.

\vspace{-0.005\textheight}
\section{Power Control Solution}\label{section: Solution}

This section presents the solution approach for the optimization problem \eqref{eq: equivalent1}, which seeks to obtain $p_t^1, \ldots, p_t^K$ that maximizes the rate-per-bit ratio $\eta_t$ at each FL iteration $t \in [T]$. The constraint \eqref{eqconstraint2} is linear in the power variables but depends on $\eta_t$ through $\theta_t^j$. Thus, at each iteration $t$, we propose solving \eqref{eq: equivalent1} using a combination of the bisection method over $\eta_t$ and linear programming~(LP) for $p_t^j$, $j = 1, \ldots, K$. This approach finds the global optimum to \eqref{eq: equivalent1} to any desired accuracy $\epsilon_B>0$. Algorithm~\ref{alg: LP + bisection} outlines the FL process with the proposed quantization scheme and power control using bisection and LP methods.

\begin{algorithm}[t]
\caption{FL over CFmMIMO with adaptive mixed-resolution quantization and reducing straggler effect by power control}
\small
\label{alg: LP + bisection}
\begin{algorithmic}[1]
\State \textbf{Inputs:} $M$, $N$, $K$, $d$, ${(\bx, y)}$, $\left\{ \bar{A_j}, \bar{B}_j, \Tilde{B}_j^{j'}, I_M^j, b_j\right\}_{j,j'} $, T,~$\bw_0$,~$\epsilon_B$
\vspace{0.001\textheight}

 \For{$t = 1, \ldots, T$,}
 {\Comment{\color{cyan}FL global iterations}}
\vspace{0.001\textheight}
\State APs receive~$\bw_{t-1}$ from the central server and send it to all users via downlink transmission

\For{$j = 1, \ldots, K$,} 
\vspace{0.002\textheight}
\State Set $\bw_{0,t}^j=\bw_{t-1} $ 
        \For{$h= 1, \ldots, L$} \Comment{{\color{cyan}Local iterations}}
            \State update $\bw_{h,t}^j$ according to~\eqref{eq: local trains}
            \vspace{0.001\textheight}     
            
        \EndFor 
        \State Set $\bw_{t}^j=\bw_{L,t}^j $ and~$\delta\bw_t^j= \bw_{t}^j - \bw_{t-1}$ 

        \State Obtain~$b_t^j$ by quantization scheme~\eqref{eq: bits}
        \Comment{{\color{cyan}Quantization}}
    
        \vspace{0.002\textheight}
        \State Send~$b_{p,t}^j = \lceil\log_2(b_t^j) \rceil$ bits to APs via uplink transmission
\EndFor 
\vspace{0.005\textheight}
 \State {Central server do:}
 {\Comment{\color{cyan}Power control}}
  \vspace{0.005\textheight}
 \State {Initialize:} $\eta_{\text{min}}$, $\eta_{\text{max}}$ 
 
  \vspace{0.002\textheight}
  
\While{$\eta_{\text{max}} - \eta_{\text{min}} > \epsilon_B$}
{\Comment{\color{cyan}bisection over~$\eta_t$}}
    \vspace{0.002\textheight}
    \State $\eta_{t,\text{mid}} \gets ({\eta_{\text{min}} + \eta_{\text{max}}})/{2}$
    \vspace{0.002\textheight}

    \State Solve~\eqref{eq: equivalent1} for $\eta_t = \eta_{t,\text{mid}}$
    {\Comment{\color{cyan}Solve the linear program}}
    \vspace{0.002\textheight}

    
    \If{~\eqref{eq: equivalent1} is \textbf{feasible},}
        \State $\eta_{\text{min}} \gets \eta_t$ {\Comment{\color{cyan}Increase lower bound}}
    \Else
        \State $\eta_{\text{max}} \gets \eta_t$ {\Comment{\color{cyan}Decrease upper bound}}
    \EndIf
\EndWhile 
\vspace{0.002\textheight}
\State Central server send~$\eta_t^* = \eta_{t,\text{mid}}$, and $\bp_t^* = \bp_t$ to users via downlink transmission
\vspace{0.002\textheight}
\For {$j=1,\ldots,K$,} in parallel:
\vspace{0.002\textheight}
\State  Send the quantized local vector with~$b_t^j$ bits to the APs via uplink transmission and uplink power~${p_t^j}^*$
\EndFor

\vspace{0.002\textheight}
\State Central server waits until it receives all~$\widehat{\delta\bw}_t^j$ and update the global model as $\bw_{t} \leftarrow \bw_{t-1} + \sum_{j =1}^{K} \rho_j~\widehat{\delta\bw}_t^j  $ 

\vspace{0.002\textheight}
\EndFor 
\State \textbf{Output:} $\bw_T$, $\bp_t^*|_{t=1,\ldots,T}$, $\eta_t^*|_{t=1,\ldots,T}$

\end{algorithmic}
\end{algorithm}
\begin{figure*}[t]
\begin{minipage}{0.99\columnwidth}
\includegraphics[width=\columnwidth] {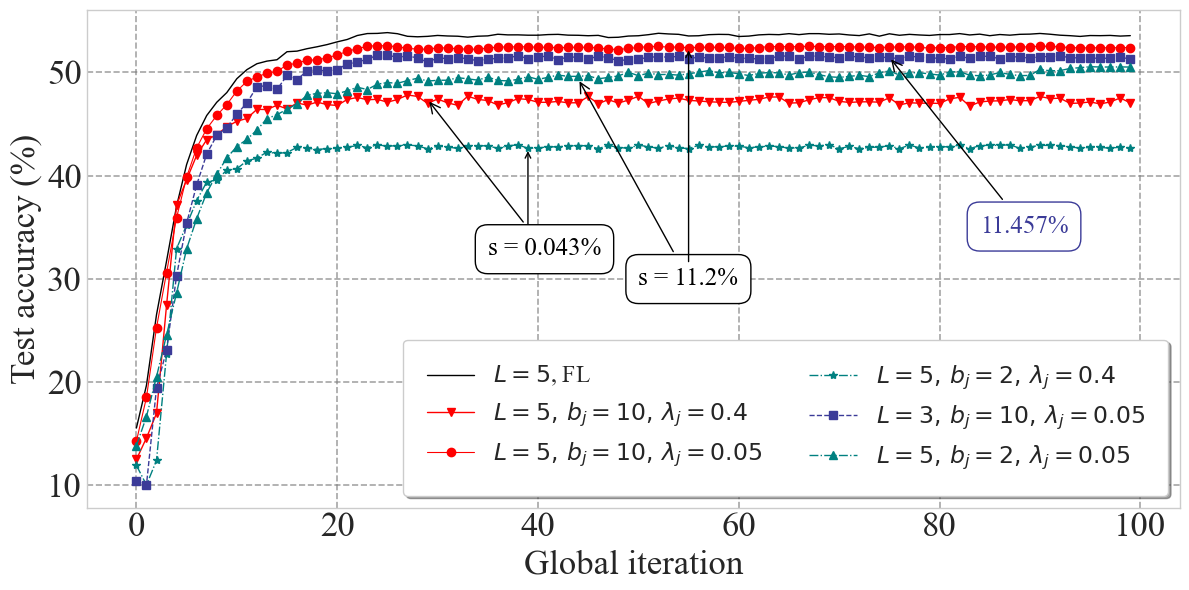}
\subcaption{$K = 40$.}
\label{subfig: K40}
\end{minipage}
\begin{minipage}{0.99\columnwidth}
\includegraphics[width=\columnwidth] 
{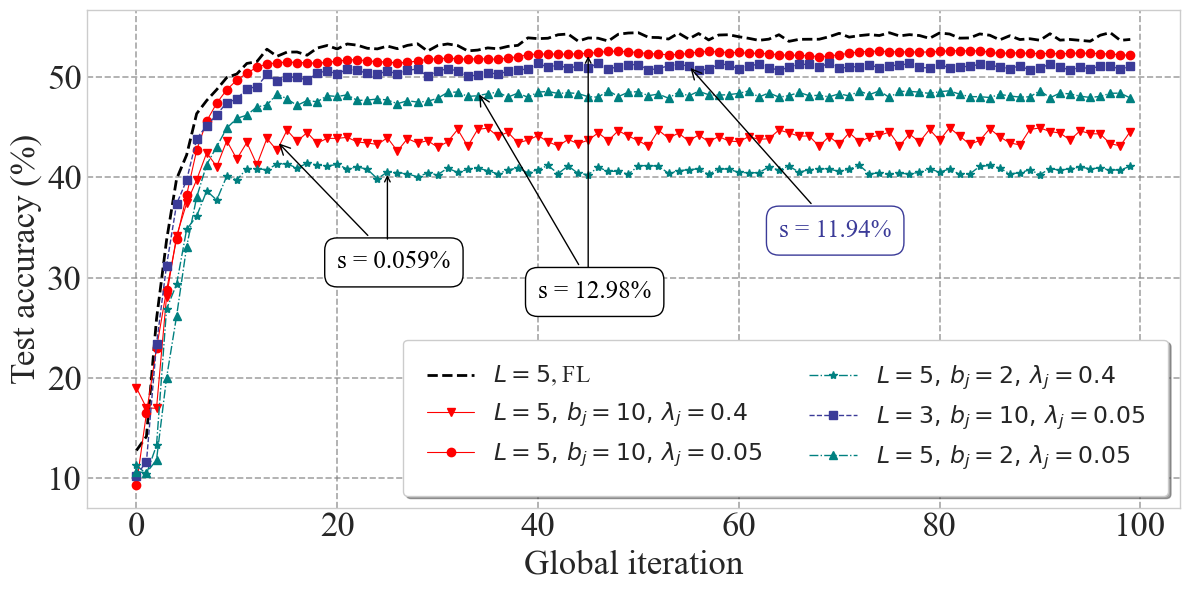}
\subcaption{$K = 20$.}
\label{subfig: K20}
\end{minipage}
\caption{Convergence analysis of FL with the mixed-resolution quantization vs. classic FL for CIFAR-10 dataset with non-IID data distribution among users. 
}\vspace{-0.012\textheight}
\label{fig: convergence_vs_FL}

\end{figure*} 
%

In each iteration $t$, the APs receive the global model $\bw_{t-1}$ and distribute it to the users (lines 2-3). Each user $j$ performs $L$ local updates to compute $\bw_{h,t}^j$ according to \eqref{eq: local trains} (line 7). The users then quantize $\delta \bw_{t}^j$ using \eqref{eq: bits}, determine $b_t^j$, and transmit $b_t^j$ with $b_{p,t}^j = \lceil\log_2(b_t^j)\rceil$ bits at full uplink power $p_{p,t}^j = 1$ (lines 10-11). As $b_t^j \le 32d$, we have $b_{p,t}^j \le \log_2(32d) \le 32$, making the latency for $b_{p,t}^j$ negligible compared to the uplink latency $\ell_t^j$. The server then solves~\eqref{eq: equivalent1} via bisection and LP (lines 13-23), providing users with the optimal powers $\bp_t^*$ and $\eta_t^*$ (line 24). The users transmit $\delta \bw_{t}^j$ using power ${p_t^j}^*$ (line 26), and the server aggregates them to update $\bw_{t}$ (line 28). This procedure repeats for~$T$ global iterations. 


\vspace{-0.012\textheight}
\section{Numerical Results}\label{section: Numericals}
\vspace{-0.005\textheight}
In this section, we demonstrate the efficiency of our proposed solution by considering a CFmMIMO network with $M=16$ APs, each with~$N=4$ antennas, and $K \in \{ 20,40\}$ single-antenna FL users. Each user is assigned to a $\tau_p$-length pilot sequence according to the assignment algorithm in~\cite{cellfreebook}. The simulation parameters and values are shown in TABLE~\ref{tab: parameters}. 
\begin{table}[t]
    \centering
    \caption{Simulation parameters.}
    \label{tab: parameters}
    \renewcommand{\arraystretch}{1} 
    \setlength{\extrarowheight}{0.9pt} 
    
    \resizebox{\columnwidth}{!}{%
    \begin{tabular}{|l l||l l|}
        \hline
        \textbf{Parameter} &  \textbf{Value} & \textbf{Parameter} &  \textbf{Value}\\
        \hline
        Bandwidth & $B = 20$\,MHz & Number of users & $K \in \{20, 40\}$\\
        \hline
        Area of interest (wrap around) & $1000 \times 1000$m & Pathloss exponent & $\alpha_p= 3.67$\\
        \hline
        Number of APs & $M = 16$ & Coherence block length & $\tau_c = 200$\\
        \hline
        Number of per-AP antennas  & $N = 4$ & Pilot length & $\tau_p = 10$\\
        \hline
        Uplink transmit power & $p^u = 100$\,mW & Uplink noise power & $\sigma^2 = -94$\,dBm\\
        \hline
        Noise figure & $7$\,dB & Number of FL local iterations & $L \in \{3, 5\}$ \\
        \hline
    \end{tabular}%
    }
    \vspace{-0.025\textheight}
\end{table}

 

The users train Convolutional Neural Networks (CNNs) on IID or non-IID local datasets. The CNN architecture is consistent across users: the input layer processes data of shape $(32, 32, 3)$ for CIFAR-10 and CIFAR-100, and $(28, 28, 3)$ for Fashion-MNIST (pre-processed from grayscale to three channels). It consists of a Conv2D layer with 32 filters of size $(3, 3)$ and ReLU activation, followed by a MaxPooling2D layer with a pool size of $(2, 2)$~\cite{mcmahan_CNN}. A Flatten layer converts the data to a 1D array for the Dense layers. The first Dense layer has 64 units with ReLU activation, and the second Dense layer has $n_2$ units with softmax activation, where $n_2 = 10$ for CIFAR-10 and Fashion-MNIST, and $n_2 = 100$ for CIFAR-100. This architecture captures spatial features for multiclass classification. Note that the CNN architecture is sufficient for our needs, as our goal is to evaluate our optimization method rather than achieve the best possible accuracy from the CNN. 

 We define the average percentage of high-resolution elements as $s:= 100 \cdot \E_{t,j}{s_t^j}$. We calculate the average reduction in communication overhead as~$\bar{r}:= 100 \cdot (1 - \E_{t,j}{b_t^j}/(b_1 d) ) \%$, where~$b_1$
 is the number of element-wise bits for the benchmarks ($32$ for classic FL, $b_1 = b_j$ for LAQ and Top-$q$). Considering the definition of~$s$, we obtain~$\bar{r}= 100 - 100/b_1 - s (b_j-1)/b_1$, in $\%$, as the reduction in communication overhead considering the proposed mixed-resolution quantization compared to the benchmarks.

Fig.~\ref{fig: convergence_vs_FL} shows the convergence behavior of FL on the CIFAR-10 dataset with a non-IID (independent and identically distributed) data distribution among users using the proposed mixed-resolution quantization. The annotation boxes display the average percentage of high-resolution elements. The results demonstrate that FL, with the proposed mixed-resolution quantization, achieves a convergence rate comparable to classic FL. Specifically, the proposed scheme maintains similar test accuracy considering~$\lambda_j = 0.05$ while reducing the communication overhead by $\bar{r} = 93 \%$, highlighting that even with a small value of~$\lambda_j$, we can greatly reduce the overhead.



 \begin{table}[t]
    \centering
    \caption{Performance analysis of FL with mixed-resolution quantization for CIFAR-10, CIFAR-100, and Fashion-MNIST datasets with non-IID and IID data distributions, $K= 20$, $L= 5 $, $b_j= 10$, $\lambda_j = 0.2$, $T = 100$. }
    \label{tab: comparison}
    \renewcommand{\arraystretch}{1.3} 
    \setlength{\extrarowheight}{1.8pt} 
   
    \resizebox{\columnwidth}{!}{%
    \fontsize{13}{12}
    
\begin{tabular}{|c|| c|c|c|| c|c|c|}

\hline
 &\multicolumn{3}{c||}{\textbf{Non-IID}}&\multicolumn{3}{c|}{\textbf{IID} }\\
\cline{2-7}
& {\large Accuracy} & {\large s} & {\large Accuracy} & {\large Accuracy} &  $s$ & {\large Accuracy}\\
& (Our approach)  &  $(\%)$ &  (FL)  & (Our approach)  &  $(\%) $  &  (FL)\\
\hline
{\textbf{CIFAR-10}} &$51.86$ & $0.8574$ & $52.18$ & $58.92$& $1.827$ & $59.64$ \\
\hline
{\textbf{CIFAR-100}} &$27.9$ & $1.064$ & $28.12$ &$32.26$ & $1.089$& $32.49$  \\
\hline
{\textbf{Fashion-MNIST}} & $89.29$ & $0.711$  & $89.65$ & $91.23$ &$0.9993$ & $91.73$     \\
\hline
\end{tabular}
 }  \vspace{-0.03\textheight}
\end{table}


To evaluate the effectiveness of the proposed mixed-resolution quantization scheme across various datasets, TABLE~\ref{tab: comparison} presents the test accuracy of FL with and without this scheme under both IID and non-IID data distributions for CIFAR-10, CIFAR-100, and Fashion-MNIST datasets. The results indicate that the proposed mixed-resolution quantization achieves nearly the same test accuracy as classic FL while reducing the communication overhead by at least $\bar{r}= 96 \%$.

We will now evaluate how the proposed Algorithm~\ref{alg: LP + bisection} mitigates the straggler effect using power control. We compare it with two benchmarks: the Dinkelbach method~\cite{EE_dinkel} and max-sum rate~\cite{cellfreebook}. For local gradient vectors, we employ our proposed mixed-resolution method alongside AQUILA~\cite{AQUILA}, LAQ~\cite{LAQ}, and Top-$q$ sparsification, where only the $q$ largest magnitude entries are transmitted~\cite{wangni2018gradient}. The total latency is calculated as the sum of uplink latency and computation time of local gradients. The maximum computation time of the users is computed as $\ell_c = L |\calD| a_j /(K \nu_j)$\cite{fedcau}, where $\nu_j = 20$ cycles/s, $a_j = 10^6$ cycles/sample, and $5 \times 10^4$ samples for the CIFAR-10 dataset distributed among users in a non-IID manner. The maximum number of iterations that each approach can perform within the total training latency budget $\bar{\ell}$ is reported as $T_{\max}$ in TABLE~\ref{tab: comparison 2}. Our results demonstrate that the proposed algorithm achieves superior test accuracy compared to AQUILA, LAQ, and Top-$q$ with $q = s$, underscoring the importance of considering signs for low-resolution elements to enhance convergence speed. As shown in TABLE~\ref{tab: comparison 2}, our power control method outperforms the Dinkelbach and max-sum rate by effectively reducing the straggler effect, resulting in a higher $T_{\max}$ under a limited latency constraint. The main columns in TABLE~\ref{tab: comparison 2} (LAQ, Top-$q$, AQUILA, mixed-resolution) assess communication overhead for each power control scheme (rows), leading to different total latencies for each communication method and highlighting the convergence speed of our approach and Algorithm~\ref{alg: LP + bisection} under a limited total latency budget. Our algorithm reduces communication overhead by $\bar{r} = 75\%$ compared to AQUILA and LAQ, showing a great ability to mitigate straggler effects and allow more iterations within a given total latency budget. Thus, the proposed adaptive mixed-resolution quantization scheme combined with the power control of Algorithm~\ref{alg: LP + bisection} are suitable approaches for latency-critical applications of FL, specifically in next-generation wireless networks.




\begin{table}[t]
    \centering
    \caption{ Performance comparison of our proposed method with the benchmarks, $K=40$, $L = 5$, $b_j = 4$,  $\lambda_j = 0.4$ for non-IID distributed CIFAR-10, with total latency budget of $\bar{\ell} = 3$ s, and calculated~$s = 0.044 \%$. }
    \label{tab: comparison 2}
    \renewcommand{\arraystretch}{1.3} 
    \setlength{\extrarowheight}{1.8pt} 

    \resizebox{\columnwidth}{!}{%
    \fontsize{13}{12}
    
    \begin{tabular}{|c||c|c||c|c||c|c||c|c|}
    \hline
     & \multicolumn{2}{c||}{\textbf{LAQ}} & \multicolumn{2}{c||}{\textbf{Top-$q$}} & \multicolumn{2}{c||}{\textbf{AQUILA}} & \multicolumn{2}{c|}{\textbf{Mixed-resolution}} \\
    \cline{2-9}
     & {\large Accuracy (\%)} & {$T_{\max}$} & {\large Accuracy (\%)} & {$T_{\max}$} & {\large Accuracy (\%)} & {$T_{\max}$} & {\large Accuracy (\%)} & {$T_{\max}$} \\
    \hline
    {\textbf{Our power control }  }  & $25.4 \%$ & $17$ & $36.34 \%$ & $38$ & $27.77 \% $& $16$ & {\boldsymbol{$46.13 \%$}}  & $27$  \\
    \hline
    {\textbf{Dinkelbach}}  & $22.23\%$ & $11$ & $33.7 \%$ & $36$ & $26.72 \%$ & $14$ & $42.8 \%$ & $24$  \\
    \hline
    {\textbf{Max-sum rate}} & $11.03 \%$ & $1$ & $12.93 \%$ & $5$ & $11.51 \%$ & $1$ & $25.09 \%$ & $4$  \\
    \hline
    \end{tabular}
    }
    \vspace{-0.026\textheight}
\end{table}

 \vspace{-0.008\textheight}
\section{Conclusions and future work}\label{section: Conclusion}
In this paper, we co-optimized the physical layer of CFmMIMO with the FL application to mitigate the straggler effect. We introduced a novel adaptive mixed-resolution quantization scheme and a dynamic uplink power control strategy. Our method prioritized essential entries in the local gradient updates and efficiently managed user rates. The results showed that our approach achieved test accuracy comparable to classic FL while reducing communication overhead by at least $93\%$ across CIFAR-10, CIFAR-100, and Fashion-MNIST datasets with both IID and non-IID distributions. Compared to the AQUILA, Top-$q$, and LAQ benchmarks, our method reduced communication overhead by $75\%$ and improved the test accuracy by $10\%$ within a constrained total latency budget, highlighting its effectiveness in latency-critical FL applications, especially for intelligent next-generation wireless networks.

For future work, we will explore using an adaptive number of element-wise bits based on the norm of the local FL gradients and design magnitude ratio thresholds tailored to each user's channel conditions to further reduce total latency. 

\vspace{-0.006\textheight}
\section*{{Appendix A: Proof of Lemma~\ref{lemma: quant error}}\label{P: lemma: quant error}}
\vspace{-0.005\textheight}
Considering that there exist two quantization resolutions for any entry of the vector~$\delta \bw_t^j$, the value of $\|\boldsymbol{\varepsilon}_t^j\|_{\infty} $ is the maximum of the following errors. According to~$\boldsymbol{\varepsilon}_t^j:= \delta \bw_t^j - \widehat{\delta\bw}_t^j$, for high-resolution elements satisfying~\eqref{eq: quant_factor} with uniform quantization by~$b_j$ bits and grid radius of~$r_t^j$, we have
\vspace{-0.02\textheight}
\begin{alignat}{3}\label{eq: err_high}
    \left| [\boldsymbol{\varepsilon}_t^j]_i \right| &\le \frac{r_t^j}{2(2^{b_j} -1)} = \frac{\nrm{\delta \bw_t^j}_{\infty}-{\delta w}_{q,t}^{j}}{2(2^{b_j} -1)} \: \\ 
 \nonumber
 &\le \frac{\nrm{\delta \bw_t^j}_{\infty} - \lambda_j \nrm{\delta \bw_t^j}_{\infty}}{2(2^{b_j} -1)} = \nrm{\delta \bw_t^j}_{\infty}\frac{1 - \lambda_j} {2(2^{b_j} -1)},
\end{alignat} 
\vspace{-0.015\textheight}
where~$\delta\bw_{q,t}^j \ge \lambda_j \|\delta\bw_t^j\|_{\infty}$. For the low-resolution elements~$i$ 
\vspace{-0.008\textheight}
\begin{alignat}{3}\label{eq: P: lemma quant reeor}
\nonumber
     &[\boldsymbol{\varepsilon}_t^j]_i =  [\delta\bw_t^j]_i - \frac{\widehat{\delta\bw}_{q,t}^j }{2} \le  [\delta\bw_t^j]_i - \frac{{\delta\bw}_{q,t}^j }{2} +  \frac{r_t^j}{4(2^{b_j} -1)} \stackrel{ \eqref{eq: err_high} }{\le} \: \\ 
 &
 \lambda_j\nrm{\delta\bw_t^j}_{\infty} -\frac{\lambda_j}{2}\nrm{\delta\bw_t^j}_{\infty} + \nrm{\delta \bw_t^j}_{\infty}\frac{1 - \lambda_j} {4(2^{b_j} -1)}. 
\end{alignat}


 \vspace{-0.01\textheight}
\bibliographystyle{./MetaFiles/IEEEtran}
\bibliography{./MetaFiles/References}
\end{document}

{\color{blue}The figures/table will be:}
\begin{itemize}
 \item Test accuracy vs $t$ for different~$b_j$ (maybe different local iterations $L$ as well if we have time to simulate)
    \item Simulating with CIFAR-10, CIFAR-100, and possibly FEMNIST to investigate the dataset's effect (the CIFAR-100 works with constant lambda as well as CIFAR-10).
    \item Test accuracy vs $t$ compared to LAQ and Top-$q$ with the same effective bits and percentage for the same power allocation (Thus, investigating the outperformance of the quantization scheme for similar costs)

    \item another benchmark can be the quantization that assigns 0 to the low-value elements and compare the convergence
    \item Test accuracy vs $t$ for the similar bits, but with max-min and the Dinkelbach power allocation schemes to highlight the outperformance of our power allocation even with the extra transmission of~$b_{p,t}^j$ at every iteration.
    (my cat's contribution: §1§1§1§1§1§1§1§1§1§1§1§1§1)                              
    
\end{itemize}

{\textbf{Assumption~1:}} (Unbiased Gradient). $\nabla_{\bw} f_j(\bw; \bx_{nj}, y_{nj})$ is an unbiased estimation of each user's true gradient~$\nabla_{\bw}f_j(\bw)$, for any~$j \in [K], n~\in~[ |\xi| ]$. 
\newline
{\textbf{Assumption~2:}} (Lipschitz Gradient). The function~$f_j$, for all $j \in [K]$ is~$\bar{L}$-smooth, $0 < \bar{L} < \infty$, i.e.,~$\nrm{\nabla f_j(\bw_1)-\nabla f_j(\bw_2)} \le \bar{L} \left\| \bw_1 - \bw_2 \right\|$, for all~$\bw_1,~\bw_2~\in~\R^d$.
\newline
{\textbf{Assumption~3:}} (Bounded Local Variance). The function~$f_j$, has~$\sigma_l$-bounded variance as $\E_{\xi}~{\nrm{ [\n f_j(\bw; \bx_{nj}, y_{nj})]_i - [\n f_j(\bw)]_i}}^2 = \sigma_{l_i}^2$, for all $j \in [K]$~$\bw~\in~\R^d$, $i~\in~[d]$, $n~\in~[ |\xi| ]$ . 
\newline
{\textbf{Assumption~4:}} (Bounded Global Variance). Considering the gradient of the global loss function~$\n_{\bw} f(\bw)$, the global variance is bounded as $(1/M)\sum_{j=1}^M \nrm{[\n f_j(\bw)]_i-[\n f(\bw)]_i}^2 \le \sigma_{g_i}^2$, for all $j \in [K]$,~$\bw~\in~\R^d$, $i~\in~[d]$. 
\newline
{\textbf{Assumption~5:}} (Bounded Gradients). The function~$f_j$ has~$G$-bounded gradients as~$\left |[\n f_j(\bw; \bx_{nj}, y_{nj})]_i \right | \le G$ for any~$j \in [K], n~\in~[\xi]$,~$\bw~\in~\R^d$ and~$i~\in~[d]$.

The bounded gradient assumption is widely used for
non-convex decentralized gradient optimization. Moreover, it holds for many neural network training tasks
due to the cross-entropy loss function~\cite{sun2022decentralized}.


\begin{algorithm}[t]
\caption{Iterative Linear Programming for Power Allocation Optimization}
\label{alg:iterative_LP}
\begin{algorithmic}[1]
\State \textbf{Inputs:}
\begin{itemize}
    \item $A_{\text{bar}}$: Vector of constants for users.
    \item $B_{\text{bar}}$: Vector of constants for users.
    \item $B_{\text{tilde}}$: Matrix of interference coefficients.
    \item $I_M$: Vector of interference from other sources.
    \item $\beta$: Vector of latency constants.
    \item $T$: Number of time slots.
    \item $K$: Number of users.
\end{itemize}

\State \textbf{Initialize:}
\begin{itemize}
    \item $\eta_{t,\text{lower}} \gets 0.0$
    \item $\eta_{t,\text{upper}} \gets 10.0$
    \item $\epsilon \gets 1e-6$
    \item $\text{max\_iter} \gets 100$  \Comment{Maximum iterations for bisection and power update}
\end{itemize}

\textbf{bisection Method for $\eta_t$:}
\begin{algorithmic}[1]
    \State $\text{best\_eta\_t} \gets \eta_{t,\text{lower}}$
    \State $\text{best\_powers} \gets \text{None}$
    \While{$\eta_{t,\text{upper}} - \eta_{t,\text{lower}} > \epsilon$}
        \State $\eta_{t,\text{mid}} \gets (\eta_{t,\text{lower}} + \eta_{t,\text{upper}}) / 2.0$
        \State Initialize $\text{powers} \gets \frac{1}{K} \cdot \mathbf{1}_{K}$ \Comment{Initial guess for power allocations}
        \State $\text{powers\_old} \gets 10 \cdot \frac{1}{K} \cdot \mathbf{1}_{K}$ \Comment{Previous iteration powers}
        \State $\text{iter\_LP} \gets 0$
        \While{$\text{iter\_LP} < \text{max\_iter}$}
            \State $\text{powers\_old} \gets \text{copy}(\text{powers})$
            \For{$j \gets 0$ to $K-1$}
                \State Compute numerator and denominator using current $\eta_t$
                \State Update $p_t^j$
            \EndFor
            \State Check feasibility of $\text{powers}$
            \If{feasible}
                \State $\text{best\_eta\_t} \gets \eta_{t,\text{mid}}$
                \State $\text{best\_powers} \gets \text{powers}$
                \State Adjust $\eta_{t,\text{lower}}$ or $\eta_{t,\text{upper}}$ based on feasibility
            \EndIf
            \State Increment $\text{iter\_LP}$
        \EndWhile
        \State Output $\text{best\_eta\_t}$ and $\text{best\_powers}$
    \EndWhile
\end{algorithmic}

\textbf{Output Results:}
\begin{itemize}
    \item Print maximum $\eta_t$ and corresponding optimal power allocations.
    \item Compute SINR and latencies based on optimal $\eta_t$ and $\text{best\_powers}$.
\end{itemize}
\end{algorithmic}

\end{algorithm}

----------------------
To implement this adaptive quantization scheme, we first determine~$\lambda_j$ for each user~$j~\in~[K]$ by a heuristic approach where users supporting higher rates can transmit more bits. We define~$\lambda_j := \omega (1  - \log_2(1+\text{SINR}_{f,j}/\max_k \log_2(1+\text{SINR}_{f,k}) ) ) + \epsilon_{0,j}$, which
\begin{equation}
  \text{SINR}_{f,j} := \frac{\bar{A}_j}{\bar{B}_j + \sum_{j' \neq j}^K \Tilde{B}_{j}^{j'} + I_M^j},  
\end{equation}
where~$\omega~\le 1$ is a positive coefficient indicating the maximum intended ratio between elements we want to consider in the quantization, $0 < \epsilon_{0,j} \ll 1$ is a small constant to prevent $\lambda_j$ from being zero, and full uplink power is assumed for all users. Subsequently, for elements~$i=1,\ldots, d$, we define

%% file: MetaFiles/packages.tex
\usepackage{gensymb}
\usepackage{dsfont}
\usepackage{amsmath,amssymb,amsfonts,amsthm}

\usepackage{epsfig}
\usepackage{cite}
\usepackage{hhline}
\usepackage{multirow}
\usepackage{xcolor}
\usepackage{makecell}
\usepackage{subcaption}
\usepackage{capt-of}
\usepackage[labelformat=simple]{subcaption}

\usepackage{enumerate}
\usepackage{enumitem}
\usepackage{color}
\usepackage{graphicx}
\usepackage{algpseudocode}
\usepackage[ruled]{algorithm}
\makeatletter
\newcounter{parentalgorithm}

\makeatother

\usepackage[nolist,printonlyused]{acronym}      
\usepackage{booktabs}
\usepackage{bm}
\usepackage{pgf}
\usepackage{pgfplots}
\usepackage{tikz}
\usepackage{grffile}
\pgfplotsset{compat=newest}
\usetikzlibrary{plotmarks}
\usetikzlibrary{shapes.geometric,quotes,angles,automata,arrows,positioning,calc,3d,matrix,decorations.markings,matrix}

\tikzstyle{startstop} = [rectangle, rounded corners, minimum width=3cm, minimum height=1cm,text centered, draw=black, fill=white!30]
\tikzstyle{io} = [trapezium, trapezium left angle=70, trapezium right angle=110, minimum width=3cm, minimum height=1cm, text centered, draw=black, fill=white!30]
\tikzstyle{process} = [rectangle, minimum width=2cm, minimum height=1cm, text centered, draw=black, fill=white!30]
\tikzstyle{decision} = [diamond, minimum width=3cm, minimum height=1cm, text centered, draw=black, fill=white!30]
\tikzstyle{arrow} = [thick,->,>=stealth]

\makeatletter
\newcommand{\vast}{\bBigg@{3}}
\newcommand{\Vast}{\bBigg@{4}}
\makeatother
\renewcommand{\IEEEQED}{\IEEEQEDopen}

\showoutput
\makeatletter
\tikzset{%
  remember picture with id/.style={%
    remember picture,
    overlay,
    save picture id=#1,
  },
  save picture id/.code={%
    \edef\pgf@temp{#1}%
    \immediate\write\pgfutil@auxout{%
      \noexpand\savepointas{\pgf@temp}{\pgfpictureid}}%
  },
  if picture id/.code args={#1#2#3}{%
    \@ifundefined{save@pt@#1}{%
      \pgfkeysalso{#3}%
    }{
      \pgfkeysalso{#2}%
    }
  }
}

\def\savepointas#1#2{%
  \expandafter\gdef\csname save@pt@#1\endcsname{#2}%
}

\def\tmk@labeldef#1,#2\@nil{%
  \def\tmk@label{#1}%
  \def\tmk@def{#2}%
}

\tikzdeclarecoordinatesystem{pic}{%
  \pgfutil@in@,{#1}%
  \ifpgfutil@in@%
    \tmk@labeldef#1\@nil
  \else
    \tmk@labeldef#1,(0pt,0pt)\@nil
  \fi
  \@ifundefined{save@pt@\tmk@label}{%
    \tikz@scan@one@point\pgfutil@firstofone\tmk@def
  }{%
  \pgfsys@getposition{\csname save@pt@\tmk@label\endcsname}\save@orig@pic%
  \pgfsys@getposition{\pgfpictureid}\save@this@pic%
  \pgf@process{\pgfpointorigin\save@this@pic}%
  \pgf@xa=\pgf@x
  \pgf@ya=\pgf@y
  \pgf@process{\pgfpointorigin\save@orig@pic}%
  \advance\pgf@x by -\pgf@xa
  \advance\pgf@y by -\pgf@ya
  }%
}

\makeatother

\usepackage{comment}
\usepackage{xpatch}
\makeatletter
\xpatchcmd{\algorithmic}{\itemsep\z@}{\itemsep=-0.25mm}{}{}
\makeatother

\usepackage{matlab-prettifier}

\usepackage{multicol}
\usepackage{multirow}

%% file: MetaFiles/commands.tex

\newcommand{\n}{\nabla}
\newcommand{\nrm}[1]{\left \| #1 \right \|}

\newcommand{\E}{\mathds{E}}
\newcommand{\inner}[1]{\left\langle #1 \right \rangle}
\newtheorem{theorem}{Theorem}
\newtheorem{defin}{Definition}
\newtheorem{prop}{Proposition}
\newtheorem{lemma}{Lemma}
\newtheorem{corollary}{Corollary}
\newtheorem{alg}{Algorithm}
\newtheorem{remark}{Remark}
\newtheorem{result}{Result}
\newtheorem{conjecture}{Conjecture}
\newtheorem{example}{Example}
\newtheorem{notations}{Notations}
\newtheorem{assumption}{Assumption}
\newcommand{\combin}[2]{\ensuremath{ \left( \ba{c} #1 \\ #2 \ea \right) }}
\newcommand{\diag}{{\mbox{diag}}}
\newcommand{\rank}{{\mbox{rank}}}
\newcommand{\dom}{{\mbox{dom{\color{white!100!black}.}}}}
\newcommand{\range}{{\mbox{range{\color{white!100!black}.}}}}
\newcommand{\image}{{\mbox{image{\color{white!100!black}.}}}}
\newcommand{\herm}{^{\mbox{\scriptsize H}}}  
\newcommand{\sherm}{^{\mbox{\tiny H}}}       
\newcommand{\tran}{^{\mbox{\scriptsize T}}}  
\newcommand{\tranIn}{^{\mbox{-\scriptsize T}}}  
\newcommand{\card}{{\mbox{\textbf{card}}}}
\newcommand{\asign}{{\mbox{$\colon\hspace{-2mm}=\hspace{1mm}$}}}
\newcommand{\ssum}[1]{\mathop{ \textstyle{\sum}}_{#1}}

\newcommand{\vbar}{\raisebox{.17ex}{\rule{.04em}{1.35ex}}}
\newcommand{\vbarind}{\raisebox{.01ex}{\rule{.04em}{1.1ex}}}
\newcommand{\D}{\ifmmode {\rm I}\hspace{-.2em}{\rm D} \else ${\rm I}\hspace{-.2em}{\rm D}$ \fi}
\newcommand{\T}{\ifmmode {\rm I}\hspace{-.2em}{\rm T} \else ${\rm I}\hspace{-.2em}{\rm T}$ \fi}
\newcommand{\B}{\ifmmode {\rm I}\hspace{-.2em}{\rm B} \else \mbox{${\rm I}\hspace{-.2em}{\rm B}$} \fi}
\newcommand{\Hil}{\ifmmode {\rm I}\hspace{-.2em}{\rm H} \else \mbox{${\rm I}\hspace{-.2em}{\rm H}$} \fi}
\newcommand{\C}{\ifmmode \hspace{.2em}\vbar\hspace{-.31em}{\rm C} \else \mbox{$\hspace{.2em}\vbar\hspace{-.31em}{\rm C}$} \fi}
\newcommand{\Cind}{\ifmmode \hspace{.2em}\vbarind\hspace{-.25em}{\rm C} \else \mbox{$\hspace{.2em}\vbarind\hspace{-.25em}{\rm C}$} \fi}
\newcommand{\Q}{\ifmmode \hspace{.2em}\vbar\hspace{-.31em}{\rm Q} \else \mbox{$\hspace{.2em}\vbar\hspace{-.31em}{\rm Q}$} \fi}
\newcommand{\Z}{\ifmmode {\rm Z}\hspace{-.28em}{\rm Z} \else ${\rm Z}\hspace{-.38em}{\rm Z}$ \fi}

\newcommand{\sgn}{\mbox {sgn}}
\newcommand{\var}{\mbox {var}}
\newcommand{\cov}{\mbox {cov}}
\renewcommand{\Re}{\mbox {Re}}
\renewcommand{\Im}{\mbox {Im}}
\newcommand{\cum}{\mbox {cum}}

\renewcommand{\vec}[1]{{\bf{#1}}}     

\newcommand{\vecsc}[1]{\mbox {\boldmath \scriptsize $#1$}}     
\newcommand{\itvec}[1]{\mbox {\boldmath $#1$}}
\newcommand{\itvecsc}[1]{\mbox {\boldmath $\scriptstyle #1$}}
\newcommand{\gvec}[1]{\mbox{\boldmath $#1$}}

\newcommand{\balpha}{\mbox {\boldmath $\alpha$}}
\newcommand{\bbeta}{\mbox {\boldmath $\beta$}}
\newcommand{\bgamma}{\mbox {\boldmath $\gamma$}}
\newcommand{\bdelta}{\mbox {\boldmath $\delta$}}
\newcommand{\bepsilon}{\mbox {\boldmath $\epsilon$}}
\newcommand{\bvarepsilon}{\mbox {\boldmath $\varepsilon$}}
\newcommand{\bzeta}{\mbox {\boldmath $\zeta$}}
\newcommand{\boldeta}{\mbox {\boldmath $\eta$}}
\newcommand{\btheta}{\mbox {\boldmath $\theta$}}
\newcommand{\bvartheta}{\mbox {\boldmath $\vartheta$}}
\newcommand{\biota}{\mbox {\boldmath $\iota$}}
\newcommand{\blambda}{\mbox {\boldmath $\lambda$}}
\newcommand{\bmu}{\mbox {\boldmath $\mu$}}
\newcommand{\bnu}{\mbox {\boldmath $\nu$}}
\newcommand{\bxi}{\mbox {\boldmath $\xi$}}
\newcommand{\bpi}{\mbox {\boldmath $\pi$}}
\newcommand{\bvarpi}{\mbox {\boldmath $\varpi$}}
\newcommand{\brho}{\mbox {\boldmath $\rho$}}
\newcommand{\bvarrho}{\mbox {\boldmath $\varrho$}}
\newcommand{\bsigma}{\mbox {\boldmath $\sigma$}}
\newcommand{\bvarsigma}{\mbox {\boldmath $\varsigma$}}
\newcommand{\btau}{\mbox {\boldmath $\tau$}}
\newcommand{\bupsilon}{\mbox {\boldmath $\upsilon$}}
\newcommand{\bphi}{\mbox {\boldmath $\phi$}}
\newcommand{\bvarphi}{\mbox {\boldmath $\varphi$}}
\newcommand{\bchi}{\mbox {\boldmath $\chi$}}
\newcommand{\bpsi}{\mbox {\boldmath $\psi$}}
\newcommand{\bomega}{\mbox {\boldmath $\omega$}}

\newcommand{\R}{\mathbb{R}}
\newcommand{\N}{\mathbb{N}}

\def\calA{{\mathcal A}}
\def\calB{{\mathcal B}}
\def\calC{{\mathcal C}}
\def\calD{{\mathcal D}}
\def\calE{{\mathcal E}}
\def\calF{{\mathcal F}}
\def\calG{{\mathcal G}}
\def\calH{{\mathcal H}}
\def\calI{{\mathcal I}}
\def\calJ{{\mathcal J}}
\def\calK{{\mathcal K}}
\def\calL{{\mathcal L}}
\def\calM{{\mathcal M}}
\def\calN{{\mathcal N}}
\def\calO{{\mathcal O}}
\def\calP{{\mathcal P}}
\def\calQ{{\mathcal Q}}
\def\calR{{\mathcal R}}
\def\calS{{\mathcal S}}
\def\calT{{\mathcal T}}
\def\calU{{\mathcal U}}
\def\calV{{\mathcal V}}
\def\calW{{\mathcal W}}
\def\calX{{\mathcal X}}
\def\calY{{\mathcal Y}}
\def\calZ{{\mathcal Z}}

\def\bA{\mbox {\boldmath $A$}}
\def\bB{\mbox {\boldmath $B$}}
\def\bC{\mbox {\boldmath $C$}}
\def\bD{\mbox {\boldmath $D$}}
\def\bE{\mbox {\boldmath $E$}}
\def\bF{\mbox {\boldmath $F$}}
\def\bG{\mbox {\boldmath $G$}}
\def\bH{\mbox {\boldmath $H$}}
\def\bI{\mbox {\boldmath $I$}}
\def\bJ{\mbox {\boldmath $J$}}
\def\bK{\mbox {\boldmath $K$}}
\def\bL{\mbox {\boldmath $L$}}
\def\bM{\mbox {\boldmath $M$}}
\def\bN{\mbox {\boldmath $N$}}
\def\bO{\mbox {\boldmath $O$}}
\def\bP{\mbox {\boldmath $P$}}
\def\bQ{\mbox {\boldmath $Q$}}
\def\bR{\mbox {\boldmath $R$}}
\def\bS{\mbox {\boldmath $S$}}
\def\bT{\mbox {\boldmath $T$}}
\def\bU{\mbox {\boldmath $U$}}
\def\bV{\mbox {\boldmath $V$}}
\def\bW{\mbox {\boldmath $W$}}
\def\bX{\mbox {\boldmath $X$}}
\def\bY{\mbox {\boldmath $Y$}}
\def\bZ{\mbox {\boldmath $Z$}}

\def\ba{\mbox {$\bf{a}$}}
\def\bb{\mbox {\boldmath $b$}}
\def\bc{\mbox {\boldmath $c$}}
\def\bd{\mbox {\boldmath $d$}}
\def\be{\mbox {\boldmath $e$}}
\def\bg{\mbox {\boldmath $g$}}
\def\bh{\mbox {\boldmath $h$}}
\def\bi{\mbox {\boldmath $i$}}
\def\bj{\mbox {\boldmath $j$}}
\def\bk{\mbox {\boldmath $k$}}
\def\bl{\mbox {\boldmath $l$}}
\def\bm{\mbox {\boldmath $m$}}
\def\bn{\mbox {\boldmath $n$}}
\def\bo{\mbox {\boldmath $o$}}
\def\bp{\mbox {\boldmath $p$}}
\def\bq{\mbox {\boldmath $q$}}
\def\br{\mbox {\boldmath $r$}}
\def\bs{\mbox {\boldmath $s$}}
\def\bt{\mbox {\boldmath $t$}}
\def\bu{\mbox {\boldmath $u$}}
\def\bv{\mbox {\boldmath $v$}}
\def\bw{\mbox {\boldmath $w$}}
\def\bx{\mbox {\boldmath $x$}}
\def\by{\mbox {\boldmath $y$}}
\def\bz{\mbox {\boldmath $z$}}

\newcommand{\snr}{\textup{SNR}}
\newcommand{\UE}{\mathrm{UE}}
\newcommand{\BS}{\mathrm{BS}}
\newcommand{\Passoc}{p_{_{I^{(1)}}}}
\newcommand{\Pintra}{p_{_{I^{(2)}}}}
\newcommand{\Pinter}{p_{_{I^{(3)}}}}

\newenvironment{Ex}
{\begin{adjustwidth}{0.04\linewidth}{0cm}
\begingroup\small
\vspace{-1.0em}
\raisebox{-.2em}{\rule{\linewidth}{0.3pt}}
\begin{example}
}
{
\end{example}
\vspace{-5mm}
\rule{\linewidth}{0.3pt}
\endgroup
\end{adjustwidth}}


\newcommand{\Hossein}[1]{{\textcolor{blue}{\emph{**Hossein: #1**}}}}
\newcommand{\Gabor}[1]{{\textcolor{cyan}{\emph{**Afsaneh: #1**}}}}
\newcommand{\Hadi}[1]{{\textcolor{red}{#1}}}
\newcommand{\gf}[1]{{\textcolor{cyan}{#1}}}
\newcommand{\REV}[1]{{\textcolor{blue}{#1}}}


\makeatletter
\let\save@mathaccent\mathaccent
\newcommand*\if@single[3]{%
 \setbox0\hbox{${\mathaccent"0362{#1}}^H$}%
  \setbox2\hbox{${\mathaccent"0362{\kern0pt#1}}^H$}%
  \ifdim\ht0=\ht2 #3\else #2\fi
  }
\newcommand*\rel@kern[1]{\kern#1\dimexpr\macc@kerna}
\newcommand*\widebar[1]{\@ifnextchar^{{\wide@bar{#1}{0}}}{\wide@bar{#1}{1}}}
\newcommand*\wide@bar[2]{\if@single{#1}{\wide@bar@{#1}{#2}{1}}{\wide@bar@{#1}{#2}{2}}}
\newcommand*\wide@bar@[3]{%
  \begingroup
  \def\mathaccent##1##2{%
    \let\mathaccent\save@mathaccent
    \if#32 \let\macc@nucleus\first@char \fi
    \setbox\z@\hbox{$\macc@style{\macc@nucleus}_{}$}%
    \setbox\tw@\hbox{$\macc@style{\macc@nucleus}{}_{}$}%
    \dimen@\wd\tw@
    \advance\dimen@-\wd\z@
    \divide\dimen@ 3
    \@tempdima\wd\tw@
    \advance\@tempdima-\scriptspace
    \divide\@tempdima 10
    \advance\dimen@-\@tempdima
    \ifdim\dimen@>\z@ \dimen@0pt\fi
    \rel@kern{0.6}\kern-\dimen@
    \if#31
      \overline{\rel@kern{-0.6}\kern\dimen@\macc@nucleus\rel@kern{0.4}\kern\dimen@}%
      \advance\dimen@0.4\dimexpr\macc@kerna
      \let\final@kern#2%
      \ifdim\dimen@<\z@ \let\final@kern1\fi
      \if\final@kern1 \kern-\dimen@\fi
    \else
      \overline{\rel@kern{-0.6}\kern\dimen@#1}%
    \fi
  }%
  \macc@depth\@ne
  \let\math@bgroup\@empty \let\math@egroup\macc@set@skewchar
  \mathsurround\z@ \frozen@everymath{\mathgroup\macc@group\relax}%
  \macc@set@skewchar\relax
  \let\mathaccentV\macc@nested@a
  \if#31
    \macc@nested@a\relax111{#1}%
  \else
    \def\gobble@till@marker##1\endmarker{}%
    \futurelet\first@char\gobble@till@marker#1\endmarker
    \ifcat\noexpand\first@char A\else
      \def\first@char{}%
    \fi
    \macc@nested@a\relax111{\first@char}%
  \fi
  \endgroup
}
\makeatother

\def\herm{\mathsf{H}}
\def\trans{\mathsf{T}}
\newcommand{\call}[1]{{\textsf{\small \textsc{#1}}}}
\newcommand{\callf}[1]{{\textsf{\footnotesize \textsc{#1}}}}

\def\argmax{\mathrm{arg}\max}
\def\argmin{\mathrm{arg}\min}
\renewcommand{\algorithmicrequire}{\textbf{Input:}}
\renewcommand{\algorithmicensure}{\textbf{Output:}}
\algdef{SE}[PROCEDURE]{Procedure}{EndProcedure}%
   [2]{\algorithmicprocedure\ \textproc{#1}\ifthenelse{\equal{#2}{}}{}{(#2)}}%
   {\algorithmicend\ \algorithmicprocedure}%
\algdef{SE}[FUNCTION]{Function}{EndFunction}%
   [2]{\algorithmicfunction\ \textproc{#1}\ifthenelse{\equal{#2}{}}{}{(#2)}}%
   {\algorithmicend\ \algorithmicfunction}%

\makeatletter
\newcommand\fs@betterruled{%
  \def\@fs@cfont{\bfseries}\let\@fs@capt\floatc@ruled
  \def\@fs@pre{\vspace*{5pt}\hrule height.8pt depth0pt \kern2pt}%
  \def\@fs@post{\kern2pt\hrule\relax}%
  \def\@fs@mid{\kern2pt\hrule\kern2pt}%
  \let\@fs@iftopcapt\iftrue}
\floatstyle{betterruled}
\restylefloat{algorithm}
\makeatother